\documentclass[sigconf, nonacm]{acmart}

\usepackage{tabularx}
\usepackage{dsfont}
\usepackage{bm}
\usepackage{balance}
\usepackage{booktabs}
\usepackage{graphicx}
\usepackage{amsmath, amsfonts}

\usepackage{multirow}
\usepackage[normalem]{ulem}

\useunder{\uline}{\ul}{}

\AtBeginDocument{%
  \providecommand\BibTeX{{%
    \normalfont B\kern-0.5em{\scshape i\kern-0.25em b}\kern-0.8em\TeX}}}

\copyrightyear{2026}
\acmYear{2026}
\setcopyright{cc}
\setcctype{by}
\acmConference[KDD '26]{Proceedings of the 32nd ACM SIGKDD Conference on Knowledge Discovery and Data Mining V.1}{August 09--13, 2026}{Jeju Island, Republic of Korea}
\acmBooktitle{Proceedings of the 32nd ACM SIGKDD Conference on Knowledge Discovery and Data Mining V.1 (KDD '26), August 09--13, 2026, Jeju Island, Republic of Korea}
\acmDOI{10.1145/3770854.3780215}
\acmISBN{979-8-4007-2258-5/2026/08}
\setcopyright{none} 
\begin{document}

\title{Incident-Guided Spatiotemporal Traffic Forecasting}
\thanks{This paper has been accepted for publication at KDD 2026.}
\author{Lixiang Fan}
\authornote{Equally contributed to this work.}
\email{lixiangfan@buaa.edu.cn}
\affiliation{
  \institution{the State Key Laboratory of Complex and Critical Software Environment, Beihang University}
  \city{Beijing}
  \country{China}
}

\author{Bohao Li}
\authornotemark[1]
\email{libh@buaa.edu.cn}
\affiliation{
  \institution{the State Key Laboratory of Complex and Critical Software Environment, Beihang University}
  \city{Beijing}
  \country{China}
}

\author{Tao Zou}
\email{zoutao@buaa.edu.cn}
\affiliation{
  \institution{the State Key Laboratory of Complex and Critical Software Environment, Beihang University}
  \city{Beijing}
  \country{China}
}

\author{Junchen Ye}
\authornote{Corresponding Author.}
\email{junchenye@buaa.edu.cn}
\affiliation{
  \institution{School of Transportation Science and Engineering, Beihang University}
  \city{Beijing}
  \country{China}
}

\author{Bowen Du}
\email{dubowen@buaa.edu.cn}
\affiliation{
  \institution{School of Transportation Science and Engineering, Beihang University}
  \city{Beijing}
  \country{China}
}

\renewcommand{\shortauthors}{Lixiang Fan, Bohao Li, Tao Zou, Junchen Ye, and Bowen Du}

\begin{abstract}
Recent years have witnessed the rapid development of deep-learning-based, graph-neural-network-based forecasting methods for modern intelligent transportation systems. However, most existing work focuses exclusively on capturing spatio-temporal dependencies from historical traffic data, while overlooking the fact that suddenly occurring transportation incidents, such as traffic accidents and adverse weather, serve as external disturbances that can substantially alter temporal patterns. We argue that this issue has become a major obstacle to modeling the dynamics of traffic systems and improving prediction accuracy, but the unpredictability of incidents makes it difficult to observe patterns from historical sequences.
To address these challenges, this paper proposes a novel framework named the Incident-Guided Spatiotemporal Graph Neural Network (\textbf{IGSTGNN}). IGSTGNN explicitly models the incident's impact through two core components: an Incident-Context Spatial Fusion (\textit{ICSF}) module to capture the initial heterogeneous spatial influence, and a Temporal Incident Impact Decay (\textit{TIID}) module to model the subsequent dynamic dissipation. To facilitate research on the spatio-temporal impact of incidents on traffic flow, a large-scale dataset is constructed and released, featuring incident records that are time-aligned with traffic time series. On this new benchmark, the proposed IGSTGNN framework is demonstrated to achieve state-of-the-art performance. Furthermore, the generalizability of the \textit{ICSF} and \textit{TIID} modules is validated by integrating them into various existing models.
\end{abstract}

\ccsdesc[500]{Computing methodologies~Artificial intelligence}
\ccsdesc[500]{Information systems~Data mining}

\keywords{Traffic prediction, Incidents, Spatio-temporal data mining}

\maketitle

\section{Introduction}
\label{sec:introduction}

In contemporary intelligent transportation systems (ITS), the accurate perception and prediction of urban traffic networks form a critical foundation for elevating intelligent city management to a new level~\cite{yuan2018hetero, han2020traffic}. By leveraging large-scale historical data, traffic forecasting endeavors to uncover the underlying evolutionary dynamics of traffic flow, thereby enabling reliable predictions of future network states~\cite{an2021igagcn}. Such capability is indispensable for facilitating proactive traffic control, ensuring rapid emergency response, and enhancing the overall efficiency and resilience of urban road networks~\cite{zhang2024hybrid, zhao2019t}.

In recent years, the dominant paradigm in this field has shifted towards data-driven approaches, particularly deep learning models designed to capture spatio-temporal dynamics. These models often abstract the traffic network as a graph structure, enabling the capture of complex spatial dependencies through graph-based neural network architectures~\cite{Yu2017Spatio-temporal, Ta2022Adaptive, Feng2023A}. For the temporal dimension, early models primarily combined Recurrent Neural Networks (RNNs) or Convolutional Neural Networks (CNNs) to model sequential dynamics~\cite{Ju2024COOL, Ali2021Exploiting, shao2022decoupled, bai2020adaptive}. Furthermore, to capture more dynamic and long-range spatio-temporal correlations, many advanced models have incorporated attention mechanisms~\cite{xu2020spatial, wang2022attention}. These models excel at capturing the intrinsic recurrent patterns in traffic flow, such as the morning and evening peaks driven by commuting regularities, and have achieved state-of-the-art performance on numerous public benchmarks.

\begin{figure}[!htbp]
    \centering
    \includegraphics[width=0.47\textwidth]{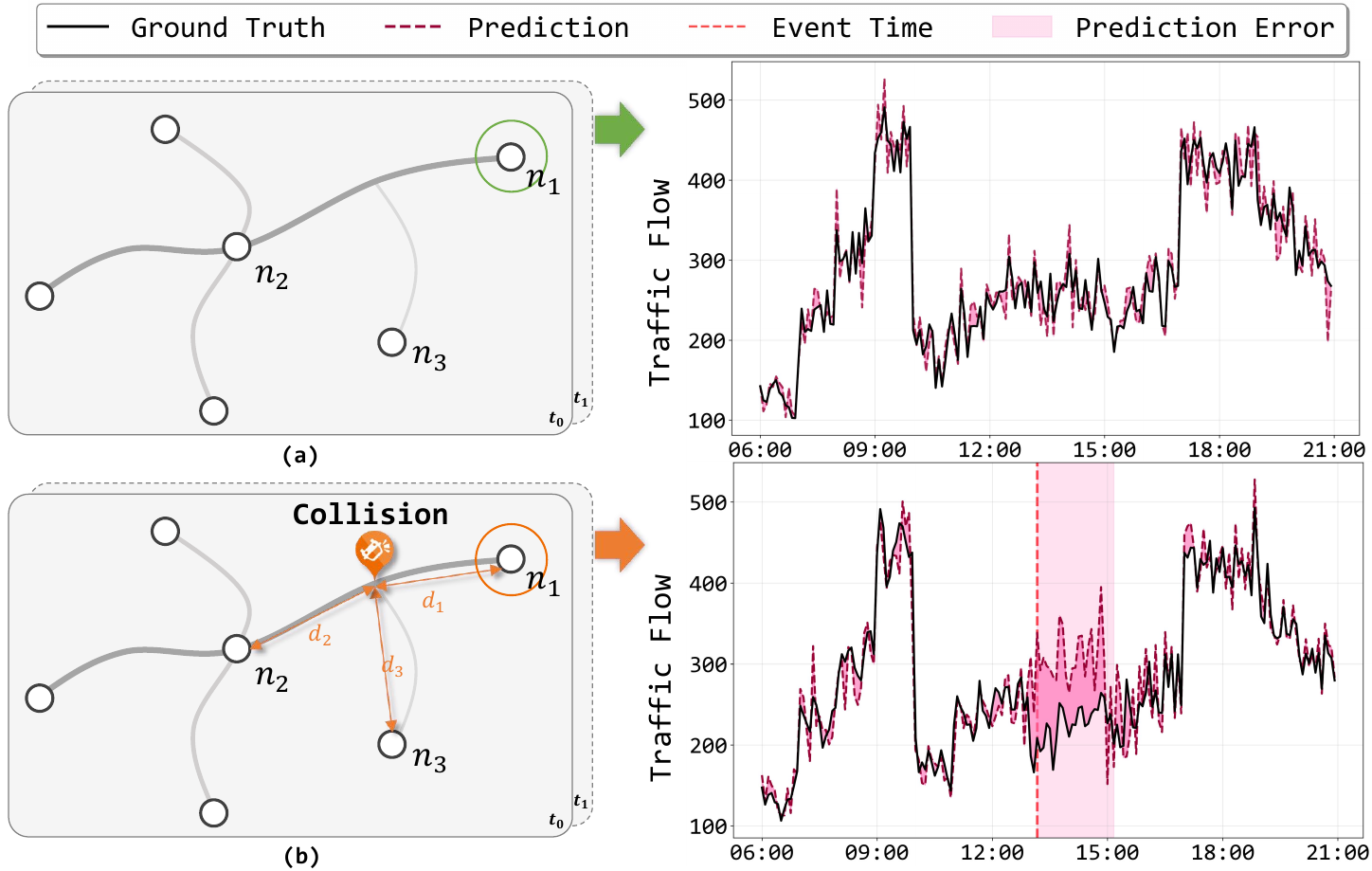}
    \caption{Illustration of a conventional model's failure under incident conditions. (a) Depicts the road network under a normal scenario, along with the typical daily traffic flow at node $n_1$. (b) Depicts the same network during an incident scenario with a collision at an intersection. The chart on the right focuses on the post-incident traffic flow at node $n_1$: the ground truth (black solid line) drops sharply, while the traditional model's forecast (dark red dashed line) completely misses this change. The pink vertical dashed line indicates the incident's start time, the pink vertical block highlights the two-hour impact duration, and the shaded red area quantifies the significant prediction error.}
    \Description{A comparison between ground truth and prediction under incident scenarios.}
    \label{fig:intro}
\end{figure}

However, real-world traffic flow is far more complex than just regular patterns, as it is constantly affected by various external disturbances such as traffic accidents, road maintenance, or adverse weather~\cite{an2021igagcn, Medina-Salgado2022Urban, du2019deep}. These external disturbances are inherently unpredictable. In historical traffic datasets, periods of regular flow are implicitly mixed with periods affected by these disturbances, often without explicit labels to differentiate them. When a model is trained solely on this mixed data, its optimization process forces it to learn a compromised function that represents an average of both normal and abnormal patterns, smoothing over the sharp dynamics caused by incidents~\cite{xu2025intervention}. This ultimately leads to significant performance degradation when the model encounters an acute incident. As illustrated in Figure~\ref{fig:intro}, a model trained on normal scenarios learns the smooth, daily traffic flow pattern for a node $n_1$ (as shown in (a)). However, when an unforeseen collision occurs at a nearby intersection (as shown in (b)), the model's forecast can only follow this learned historical average. This renders it incapable of capturing the sharp, real-world drop in traffic flow, leading to a massive prediction error. Among these various disturbances, non-recurrent incidents are a primary and particularly challenging subclass, and thus form the focus of this paper.

To effectively integrate information from such incidents and model their impact within traffic forecasting, three major challenges must be addressed.
(1) \textbf{Non-uniform Incident Impact on Spatial Domain}. Incidents could occur at any location in the road network. Even if two incidents occur between the same pair of traffic sensors, the sensors may still perceive different impacts due to the varying distances between the accident points and the sensors~\cite{Ye2023Dynamic, tran2023MSGNN}.
(2) \textbf{Dynamic Temporal Evolution of Impact}. The influence of an incident is not a singular, static impulse but a dynamic process that evolves over time~\cite{Al-Thani_Sheng_Cao_Yang_2024}. Failing to model this temporal decay is a primary cause of severe long-term prediction errors.
(3) \textbf{Diversity and Complexity of Incident Information}. The information associated with traffic incidents is characterized by both high diversity and complexity. Incidents are diverse in their nature (e.g., a multi-vehicle collision versus a stalled car), and each type induces a unique impact on traffic flow. Compounding this, the data describing any single incident is itself complex, often presented in multi-modal and unstructured formats.

To address the aforementioned challenges, we propose a novel framework named the Incident-Guided Spatiotemporal Graph Neural Network (IGSTGNN). Our framework builds upon a traditional spatio-temporal modeling backbone by introducing two core modules to solve the problem of traffic forecasting under the influence of disruptions caused by non-recurrent incidents: an Incident-Context Spatial Fusion (\textit{ICSF}) Module and a Temporal Incident Impact Decay (\textit{TIID}) Module. First, the model encodes the various inputs, including: the traffic time series, the contextual features of incidents, and the static meta-features of the sensors. Next, to achieve a fine-grained spatial fusion of the disruption, the \textit{ICSF} module employs an attention mechanism that dynamically evaluates the relationships between the incident, the regional characteristics of its location, and the real-time traffic state, generating a unique, incident-aware initial state representation for each network node. Subsequently, a spatio-temporal modeling module processes this incident-aware representation to simulate its propagation through the network. Finally, to model the dynamic temporal evolution of the incident, the \textit{TIID} module explicitly models the physical process of the incident's diminishing impact over time and superimposes it onto the base traffic evolution trend, thereby generating more accurate and reliable long-term forecasts. The main contributions of this paper are three-fold:

\begin{itemize}

    \item A novel incident-guided spatio-temporal forecasting framework, IGSTGNN, is proposed. Instead of solely learning from historical traffic patterns, this framework is designed to explicitly model the distinct spatio-temporal dynamics induced by external incidents.
    
    \item Two core components, \textit{ICSF} and \textit{TIID}, are proposed to model the incident's impact from both spatial and temporal dimensions. The \textit{ICSF} module is responsible for capturing the initial, non-uniform spatial influence, while the \textit{TIID} module models the subsequent dynamic dissipation of this impact over time.

    \item A comprehensive, real-world dataset for incident-guided forecasting is constructed and released to the public. On this benchmark, the proposed IGSTGNN framework is shown to achieve state-of-the-art performance. Moreover, the generalizability and effectiveness of the \textit{ICSF} and \textit{TIID} modules are further demonstrated by the improvement of integrating them into various existing models.
\end{itemize}

\section{Preliminaries}
\label{sec:preliminaries}

In this section, we formally define the fundamental data structures and formulate the task of incident-aware traffic forecasting. 

\textbf{Definition 1. Traffic Network and Spatio-Temporal Time Series.}
The road network is represented as a weighted undirected graph $\mathcal{G} = (\mathcal{V}, \mathcal{E})$, where $\mathcal{V} = \{v_1, \ldots, v_N\}$ is the set of $N$ traffic sensors and $\mathcal{E}$ is the set of edges representing road connections. The spatial relationships between sensor nodes are described by the weighted adjacency matrix $\bm{A} \in \mathbb{R}^{N \times N}$. The dynamic traffic conditions on this network are represented by a feature tensor $\bm{X} \in \mathbb{R}^{T \times N \times C}$, where a slice $\bm{x}_t \in \mathbb{R}^{N \times C}$ captures the state of all $N$ sensors at time step $t$.

\textbf{Definition 2. Contextual Features.}
For incident-aware forecasting, we incorporate two heterogeneous sources of contextual information: static sensor attributes and dynamic incident features.
For each sensor $v_i \in \mathcal{V}$, its features are divided into two parts: semantic attributes $\bm{s}_{i,f}$ that directly describe the road segment for feature encoding (e.g., road type, lane width) and spatial attributes $\bm{s}_{i,s}$ that are used to determine spatial relationships and construct the network topology (e.g., latitude, longitude, Fwy, Abs PM). The raw features for all sensors are consolidated in a matrix $\bm{S}_{\text{raw}}$.
For each dynamic incident $e_k$, its features are subdivided into three parts: (1) intrinsic attributes $\bm{e}_{k, \text{i}}$ that describe the event itself (e.g., incident type and description); (2) auxiliary contextual attributes $\bm{e}_{k, \text{c}}$ (e.g., the incident's relative position in the input sequence and holiday status); and (3) spatial attributes $\bm{e}_{k, \text{s}}$ used for spatial anchoring (e.g., latitude, longitude, Fwy, and Abs PM). The raw features for all incidents, encompassing these three categories, are consolidated in the matrix $\bm{E}_{\text{raw}}$.

\textbf{Definition 3. Incident-Sensor Spatial Relationship Tensor.}
The spatial relationship between incidents and sensors is captured in a pre-defined tensor $\bm{D} \in \mathbb{R}^{M \times N \times 3}$. To comprehensively represent this relationship, we model it from three complementary perspectives. First, we consider two distinct distance metrics: \textbf{road network distance} is used to capture the primary, linear propagation of impact along direct road connections, while \textbf{Euclidean distance} models the secondary, "spillover" effect on geographically proximate areas (e.g., adjacent surface streets). Second, because an incident's impact is highly asymmetric and direction-dependent, the \textbf{upstream/downstream relationship} is included as a critical feature to differentiate the severe effects on upstream traffic from the often negligible effects on downstream traffic.
For each incident-sensor pair $(e_k, v_j)$, a corresponding 3-dimensional vector $\bm{D}_{kj,:}$ is constructed. Its first two components are proximity scores derived by applying a Gaussian kernel to the \textbf{Euclidean} and \textbf{road network distances}, respectively, while the third component is a binary indicator of their relative position (\textbf{1} if the incident's Absolute Postmile is greater, \textbf{0} otherwise). The tensor $\bm{D}$ thus encapsulates the multi-faceted and direction-dependent spatial relationship between incidents and sensors.

\textbf{Definition 4. Problem Statement.}
The task is formulated based on a key assumption: the impact of incidents that occurred long ago is already implicitly captured within the recent historical traffic data. However, an incident that occurs at the most recent time step, $t$, presents a unique challenge, as its full impact has not yet propagated through the network and is not reflected in the historical sequence. Therefore, our task specifically focuses on predicting future traffic states by explicitly modeling the impact of these newly occurring incidents.
Formally, let $\bm{X} = (\bm{x}_{t-T_h+1}, \ldots, \bm{x}_t)$ denote the historical traffic states and $\hat{\bm{Y}} = (\hat{\bm{y}}_{t+1}, \ldots, \hat{\bm{y}}_{t+T_p})$ denote the target prediction sequence. Given an input instance comprising the set $\{\bm{X}, \bm{S}_{\text{raw}}, \bm{E}_{\text{raw}}, \bm{D}\}$, which includes the historical traffic states, static sensor attributes, and information for new incidents occurring at time $t$, the objective is to predict the future traffic states $\hat{\bm{Y}}$ for the next $T_p$ time steps. This task can be formulated as learning a mapping function $f$:
\begin{equation}
\hat{\bm{Y}} = f(\bm{X}, \bm{S}_{\text{raw}}, \bm{E}_{\text{raw}}, \bm{D} ; \mathcal{G}, \Theta),
\end{equation}
where $\Theta$ represents the learnable parameters of the model.

\section{Methodology}
\label{sec:methodology}

\begin{figure*}[!htbp]
    \centering
    \includegraphics[width=0.95\textwidth]{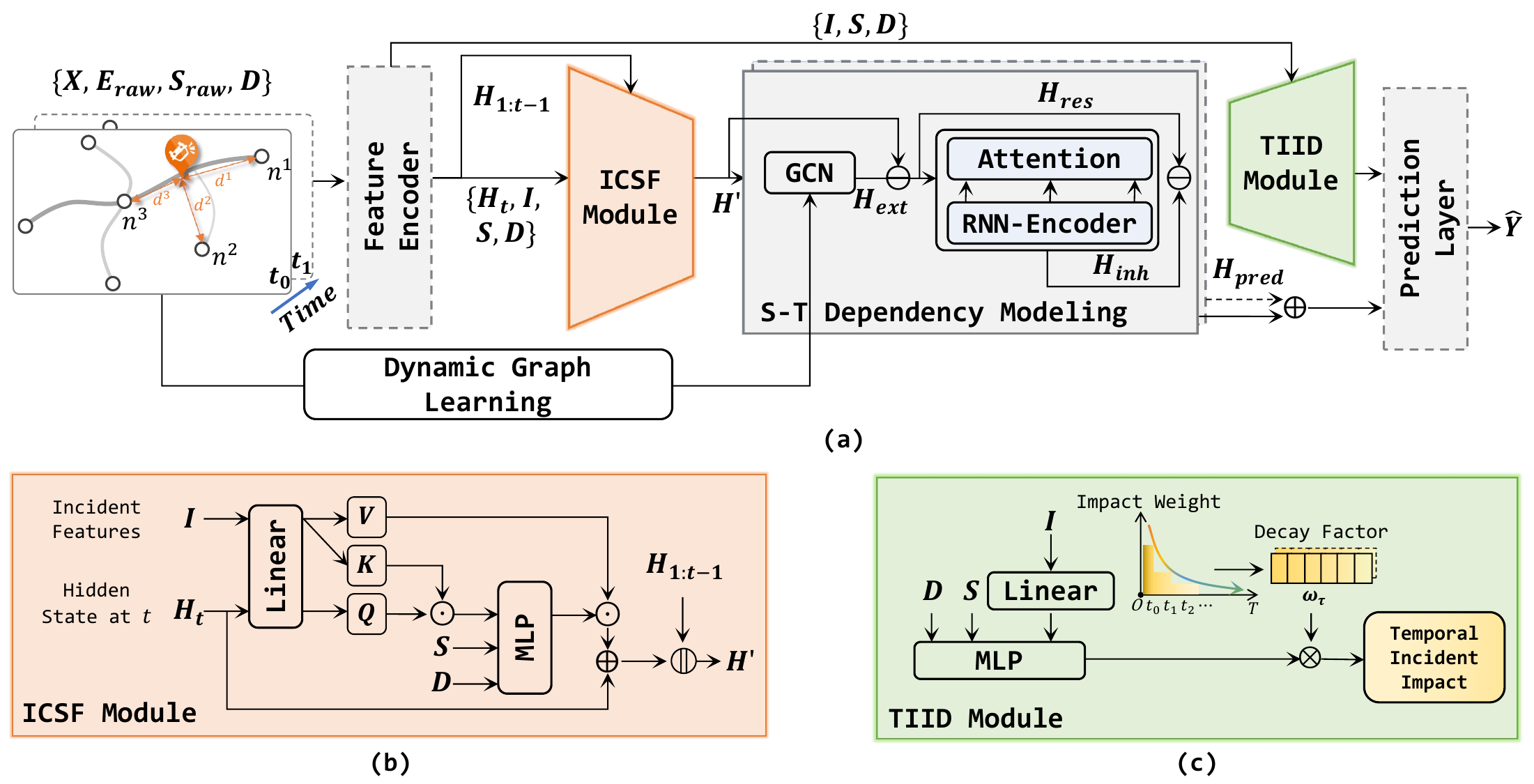}
    \caption{The overall architecture of our proposed IGSTGNN framework. (a) illustrates the main pipeline of the model, while (b) and (c) provide detailed views of the \textit{ICSF} and \textit{TIID} modules, respectively.}
    \label{fig:framework}
    \Description{
    A schematic diagram representing the IGSTGNN framework architecture. 
    Panel (a) shows the overall main pipeline, processing input traffic and incident data through multiple layers to generate predictions. 
    Panel (b) expands on the internal structure of the Incident-Context Spatial Fusion (\textit{ICSF}) module. 
    Panel (c) details the components of the Temporal Incident Impact Decay (\textit{TIID}) module.
    }
\end{figure*}

Our proposed framework, named the Incident-Guided Spatiotemporal Graph Neural Network (IGSTGNN), is designed to accurately forecast traffic conditions by explicitly modeling the influence of non-recurrent incidents. As illustrated in Figure~\ref{fig:framework}, the architecture follows a three-stage pipeline: (1) an Incident-Context Spatial Fusion (\textit{ICSF}) Module to inject heterogeneous contextual information into the traffic representation; (2) a spatio-temporal (ST) modeling module to capture the propagation and evolution of incident shocks within the network; and (3) a Temporal Incident Impact Decay (\textit{TIID}) Module to model the long-term decaying effect of the incident's impact and generate the final forecast.

\subsection{Feature Encoder}
The semantic attributes of the sensors ($\bm{s}_{i,f}$) provide rich contextual information about the road network, which is highly correlated with incident occurrence and impact~\cite{gou2024xtraffic}. Furthermore, an incident's intrinsic ($\bm{e}_{k,i}$) and contextual ($\bm{e}_{k,c}$) attributes are indispensable for the model to perceive the specific nature of each disruption.

Let $\bm{S}_{f}$ and $\bm{E}_{ic}$ denote the matrices consolidating these specific non-spatial features for all sensors and incidents, respectively. We design dedicated encoding functions, $\phi_S$ and $\phi_I$, to transform these feature matrices into dense vector representations:
\begin{equation}
    \bm{S} = \phi_S(\bm{S}_{f}), \quad \bm{I} = \phi_I(\bm{E}_{ic}),
\end{equation}
where $\bm{S} \in \mathbb{R}^{N \times d_s}$ and $\bm{I} \in \mathbb{R}^{M \times d_e}$ are the resulting encoded representations for the sensors and incidents, respectively. Concurrently, the raw traffic time series $\bm{X}$ is also projected into a high-dimensional hidden space via a linear layer to obtain its initial hidden state representation $\bm{H} \in \mathbb{R}^{T_h \times N \times d_h}$.

\subsection{Incident-Context Spatial Fusion}

An incident's impact is not uniform across the network since it is significantly dependent on factors like the distance from the event and the characteristics of the affected area. To account for this spatially heterogeneous impact, the module quantifies the influence of each incident on every sensor, thereby creating a traffic representation that is contextually aware of the event.

As depicted in Figure~\ref{fig:framework}(b), the \textit{ICSF} module is designed to capture the initial, non-uniform spatial impact of an incident. By fusing the encoded incident features ($\bm{I} \in \mathbb{R}^{M \times d_e}$) and sensor attributes ($\bm{S} \in \mathbb{R}^{N \times d_s}$) with the most recent traffic state ($\bm{H}_t \in \mathbb{R}^{N \times d_h}$), guided by the pre-defined spatial relationship tensor ($\bm{D} \in \mathbb{R}^{M \times N \times 3}$), the module produces an updated hidden state representation, $\bm{H}'_t$, which is now conditioned on the incident's context.

\textbf{Attention Component Projection.} The core of the fusion process is a scaled dot-product attention mechanism. To facilitate this, we linearly project the traffic hidden states $\bm{H}_t$ and the encoded incident representations $\bm{I}$ into three distinct subspaces:
\begin{equation}
    \bm{Q} = \bm{H}_{t} \bm{W}_Q, \quad \bm{K} = \bm{I} \bm{W}_K, \quad \bm{V} = \bm{I} \bm{W}_V,
    \label{eq:qkv_pro}
\end{equation}
where $\bm{W}_Q \in \mathbb{R}^{d_h \times d_k}$, $\bm{W}_K \in \mathbb{R}^{d_e \times d_k}$, and $\bm{W}_V \in \mathbb{R}^{d_e \times d_v}$ are learnable projection matrices. The resulting $\bm{Q} \in \mathbb{R}^{N \times d_k}$ represents the current state of each sensor, $\bm{K} \in \mathbb{R}^{M \times d_k}$ represents the properties of the incidents for matching, and $\bm{V} \in \mathbb{R}^{M \times d_v}$ contains the rich information to be aggregated from the incidents.

\textbf{Contextual Information Fusion via Attention.}
The central goal of this process is to compute a final attention weight matrix $\bm{\alpha}$ that reflects not only the interplay between traffic states and incident semantics but also incorporates prior knowledge from the road network topology.

First, we compute the initial semantic relevance score matrix via scaled dot-product attention:
\begin{equation}
    \bm{A}_{\text{sem}} = \frac{\bm{K}\bm{Q}^T}{\sqrt{d_k}},
\end{equation}
where $\bm{A}_{\text{sem}}$ is the resulting score matrix, and each element $(\bm{A}_{\text{sem}})_{kj}$ quantifies the semantic affinity between the $k$-th incident and the hidden state of the $j$-th sensor.

However, $\bm{A}_{\text{sem}}$ is based purely on feature space similarity and does not account for topological constraints of the road network. An incident should not directly influence a distant node with which it has no topological connection. To enforce this strong prior, we define a masking function, $\text{Mask}(\cdot)$, based on the spatial relationship tensor $\bm{D}$. It operates on each element of $\bm{A}_{\text{sem}}$ as follows:
\begin{equation}
    (\bm{A}'_{\text{sem}})_{kj} = 
    \begin{cases} 
        (\bm{A}_{\text{sem}})_{kj} & \text{if } (e_k, v_j) \text{ are connected in } \bm{D} \\ 
        -\infty & \text{otherwise} 
    \end{cases}.
\end{equation}

Subsequently, the masked score matrix, $\bm{A}'_{\text{sem}}$, 
is normalized via the softmax function over the incident dimension for each sensor node to obtain the preliminary attention weights:
\begin{equation}
    \tilde{\bm{\alpha}} = \text{softmax}(\bm{A}'_{\text{sem}}),
\end{equation}
where $\tilde{\bm{\alpha}}$ is the resulting preliminary, spatially-aware weights.
 
While the preliminary weights, $\tilde{\bm{\alpha}}$, incorporate topological information, they do not yet consider the unique characteristics of each sensor's location. Therefore, the preliminary weights $\tilde{\bm{\alpha}}$, the broadcasted sensor features $\bm{S}$, and the spatial relationship tensor $\bm{D}$ are concatenated and processed by a fusion function $g_{\alpha}$ and a final softmax layer (also applied over the incident dimension) to generate the final attention scores:
\begin{equation}
    \bm{\alpha} = \text{softmax}(g_{\alpha}(\tilde{\bm{\alpha}} \mathbin{\|} \bm{S} \mathbin{\|} \bm{D})),
\end{equation}
where $\bm{\alpha}$ are the final context-aware attention weights, $g_{\alpha}$ is a fusion function implemented as an MLP, and $\mathbin{\|}$ denotes the concatenation operation (with inputs broadcasted to compatible shapes).

\textbf{Weighted Aggregation.} The final attention weights $\bm{\alpha}$ are used to perform a weighted sum over the incident value representations $\bm{V}$, creating the incident context vector $\bm{C}$ for all sensor nodes:
\begin{equation}
    \bm{C} = \bm{\alpha}^{\top}\bm{V},
\end{equation}
where $\bm{C} \in \mathbb{R}^{N \times d_v}$ condenses the most relevant incident information for each sensor.

\textbf{Residual Fusion.}
Finally, the computed incident context $\bm{C}$ is added to the traffic hidden state $\bm{H}_t$ via a residual connection, and the result is normalized to produce the updated representation $\bm{H}'_{t}$:
\begin{equation}
    \bm{H}'_{t} = \text{LayerNorm}(\bm{H}_{t} + \bm{C}),
\end{equation}
where $\bm{H}'_{t}$ serves as the final, incident-aware hidden state for the subsequent spatio-temporal modeling blocks.

\subsection{Spatio-Temporal Dependency Modeling}

After the \textit{ICSF} module injects the initial impact of an incident at the final time step, the ST modeling module is tasked with simulating how this disruption propagates through the network over space and time. To effectively model this complex process, our ST module adopts a decoupling principle, aiming to separately capture two distinct traffic patterns: the external influence propagated from neighboring nodes, and the node's own inherent temporal evolution. The module processes the incident-infused sequence $\bm{H}' \in \mathbb{R}^{T_h \times N \times d_h}$ iteratively through stacked decoupling blocks.

At the core of each decoupling block is a multi-graph convolution operation, designed to capture rich spatial dependencies from heterogeneous adjacency relationships. We leverage three distinct adjacency matrices simultaneously: (1) The pre-defined static adjacency matrix $\bm{A}$, which describes the fixed physical road network structure; (2) An adaptive adjacency matrix $\bm{A}_{\text{ada}}$, constructed from learnable node embeddings to capture latent static spatial dependencies as
\begin{equation}
    \bm{A}_{\text{ada}} = \text{softmax}(\text{ReLU}(\bm{E}_u \bm{E}_d^T)),
\end{equation}
where $\bm{E}_u, \bm{E}_d \in \mathbb{R}^{N \times d_{emb}}$ denote the learnable node embedding matrices; and (3) A dynamic adjacency matrix $\bm{A}_{\text{dyn}}$~\cite{shao2022decoupled}, generated to capture transient spatial correlations that evolve over time:
\begin{equation}
    \bm{A}_{\text{dyn}} = \text{softmax}\left( \frac{(\bm{E}_{\text{dyn}} \bm{W}^{\text{dyn}}_Q)(\bm{E}_{\text{dyn}} \bm{W}^{\text{dyn}}_K)^T}{\sqrt{d_{\text{dyn}}}} \right),
\end{equation}
where $\bm{E}_{\text{dyn}} \in \mathbb{R}^{N \times d_{\text{dyn}}}$ represents the dynamic features (a concatenation of current traffic states and temporal embeddings), and $\bm{W}^{\text{dyn}}_Q, \bm{W}^{\text{dyn}}_K \in \mathbb{R}^{d_{\text{dyn}} \times d_{\text{dyn}}}$ are learnable projection matrices specific to this module.

In the $l$-th decoupling block, the input $\bm{H}^{(l-1)}$ is first processed by an \textbf{External Influence Component} using multi-graph convolution ($g_S$) to capture propagated spatial influences, yielding $\bm{H}_{\text{ext}}^{(l)}$. The inherent information flow of the nodes is then isolated via a residual connection:
\begin{equation}
    \bm{H}_{\text{res}}^{(l)} = \bm{H}^{(l-1)} - \bm{H}_{\text{ext}}^{(l)},
    \label{eq:residual_calc}
\end{equation}
where $\bm{H}_{\text{res}}^{(l)}$ represents the residual component of the hidden state, intended to capture each node's inherent trend.

Next, an \textbf{Inherent Trend Component} employs a combination of a Recurrent Neural Network (RNN) and self-attention to process the residual representation $\bm{H}_{\text{res}}^{(l)}$ and produce the refined trend component $\bm{H}_{\text{inh}}^{(l)}$. To obtain the final output of the $l$-th block, $\bm{H}^{(l)}$, this refined trend is then subtracted from the original residual, $\bm{H}_{\text{res}}^{(l)}$, in a manner analogous to the residual calculation in Equation~\eqref{eq:residual_calc}. This final output, $\bm{H}^{(l)}$, subsequently serves as the input for the $(l+1)$-th layer.

After $L$ iterations, we accumulate the forecast parts generated by the two components from each layer to obtain the final Modeled Hidden State, $\bm{H}_{\text{pred}}$. This output is a high-level representation over the prediction horizon that captures the incident-conditioned spatio-temporal dynamics and provides a robust foundation for the subsequent \textit{TIID} module to refine predictions via explicit decay modeling.

\subsection{Temporal Incident Impact Decay}

A key characteristic of traffic incidents is the dynamic nature of their impact, which evolves over time and naturally dissipates as the network recovers. Conventional models often fail to capture this temporal evolution. Therefore, our \textit{TIID} module is specifically introduced to model this decay process, simulating the dissipation of an incident's influence throughout the forecast horizon.

As shown in Figure~\ref{fig:framework}(c), the module receives the base forecast's hidden states $\bm{H}_{\text{pred}} \in \mathbb{R}^{T_p \times N \times d_{out}}$ from the ST module. To model the incident's decay effect, the module first computes the incident's initial spatial impact at the moment of occurrence, time $t$, which we term the initial incident context $\bm{C}_{\text{init}}$.

This computation aims to distill the potential impact of the incident on each node. Accordingly, the incident Key representation $\bm{K}$ is first expanded and spatially masked by the relationship tensor $\bm{D}$ to produce a localized incident representation, $\bm{K}_{\text{context}}$. This step ensures an incident only affects its spatially connected nodes. This localized representation is then concatenated with the sensor features $\bm{S}$ (dimensionally aligned via broadcasting) and the spatial relationship tensor $\bm{D}$ itself. Finally, this fused tensor is processed by a function $g_c$ to generate the initial incident context:
\begin{equation}
    \bm{C}_{\text{init}} = g_c(\bm{K}_{\text{context}} \mathbin{\|} \bm{S} \mathbin{\|} \bm{D}),
\end{equation}
where $g_c$ is a fusion function implemented as an MLP and $\mathbin{\|}$ denotes the concatenation operation.

The \textit{TIID} module employs a temporal decay factor $\omega_{\tau}$, which is modeled by a Gaussian function~\cite{Ghosh2021Comparison}. For each future prediction step $\tau \in \{1, \ldots, T_p\}$, the decay factor is calculated based on its temporal distance from the incident's occurrence:
\begin{equation}
\omega_{\tau} = \exp\left(-\frac{\tau^2}{2\sigma_t^2}\right),
\end{equation}
where $\sigma_t$ is a hyperparameter that controls the temporal decay rate. The decay factors for all time steps are combined into a vector $\bm{\omega} \in \mathbb{R}^{T_p}$. 
 The initial incident context $\bm{C}_{\text{init}}$ is then linearly transformed and modulated by this decay vector via broadcast multiplication, yielding the final Temporal Incident Impact tensor $\bm{C}_{\text{temp}}$:
\begin{equation}
\bm{C}_{\text{temp}} = \bm{\omega} \otimes (\bm{C}_{\text{init}}\bm{W}_{c}),
\end{equation}
where $\otimes$ denotes broadcasted multiplication.

Finally, the temporal incident impact tensor $\bm{C}_{\text{temp}}$ is integrated with the base forecast from the spatio-temporal backbone, $\bm{H}_{\text{pred}}$, to generate the final prediction tensor $\hat{\bm{Y}}$:
\begin{equation}
\hat{\bm{Y}} = g_{\text{out}}(\bm{H}_{\text{pred}} + \bm{C}_{\text{temp}}),
\end{equation}
where \(g_{\text{out}}\) is a multi-layer perceptron serving as the prediction head.

\section{Experiments}
\label{sec:experiments}

In this section, we experimentally validate the effectiveness of the proposed IGSTGNN framework. Following an introduction to the experimental setup (datasets, baselines, and metrics), we benchmark IGSTGNN against state-of-the-art models. The generalizability and design superiority of the core \textit{ICSF} and \textit{TIID} modules are then verified, followed by a detailed ablation study to assess the contribution of each component.

\subsection{Experimental Settings}
This subsection details our experimental setup, encompassing four key aspects: the datasets, the baseline models, the implementation details, and the metrics.

\begin{table}[b]
\centering
\caption{Statistics of the experimental datasets.}
\label{tab:dataset_stats}
\begin{tabular}{l | c | c | c }
\toprule
\textbf{Dataset} & \textbf{\# Nodes} & \textbf{\# Edges}  & \textbf{\# Incidents} \\
\midrule
Alameda & 521 & 13,828 & 14,687 \\
Contra Costa & 496 & 13,339  & 5,587 \\
Orange & 990 & 29,142 & 18,700 \\
\bottomrule
\end{tabular}
\end{table}

\begin{table}[!htbp]
\centering
\caption{Description of Sensor Meta-features.}
\label{tab:sensor_features}
\begin{tabularx}{\columnwidth}{c l l >{\raggedright\arraybackslash}X}
\toprule
\textbf{Symbol} & \textbf{Feature Name} & \textbf{Type} & \textbf{Example} \\
\midrule
$s_1$ & Type & String & Mainline \\
$s_2$ & Surface & String & Asphalt \\
$s_3$ & Roadway Use & String & Commercial \\
$s_4$ & Lane Width & Float & 3.7 \\
$s_5$ & Design Speed Limit & Integer & 105 \\
$s_6, s_7$ & Latitude, Longitude & Float & 34.0522, -118.2437 \\
$s_8$ & Freeway Name & String & I-405 \\
$s_9$ & Absolute Postmile & Float & 23.15 \\
\bottomrule
\end{tabularx}
\end{table}

\begin{table}[!htbp]
\centering
\caption{Description of Incident Information Features.}
\label{tab:incident_features}
\begin{tabularx}{\columnwidth}{c l l >{\raggedright\arraybackslash}X}
\toprule
\textbf{Symbol} & \textbf{Feature Name} & \textbf{Type} & \textbf{Example} \\
\midrule
$e_1$ & Relative Position & Integer & 11 \\
$e_2$ & Description & String & Traffic Collision \\
$e_3$ & Type & String & Accident \\
$e_4$ & Holiday & Cat. & 0 \\
$e_5, e_6$ & Latitude, Longitude & Float & 34.0592, -118.4452 \\
$e_7$ & Absolute Postmile & Float & 25.50 \\
$e_8$ & Freeway Name & String & I-405 \\
\bottomrule
\end{tabularx}
\end{table}

\begin{table*}[!htbp]
\centering
\caption{Performance comparison on the Alameda, Contra Costa, and Orange datasets. The best results are in \textbf{bold} and the second-best results are \underline{underlined}.}
\label{tab:main_results_all_datasets}
\resizebox{\textwidth}{!}
{
\begin{tabular}{l l|ccc|ccc|ccc|ccc}
\toprule
\multirow{2}{*}{Data} & \multirow{2}{*}{Method} & \multicolumn{3}{c|}{Horizon 3} & \multicolumn{3}{c|}{Horizon 6} & \multicolumn{3}{c|}{Horizon 12} & \multicolumn{3}{c}{Average} \\ \cline{3-14} 
 &  & MAE & RMSE & MAPE(\%) & MAE & RMSE & MAPE(\%) & MAE & RMSE & MAPE(\%) & MAE & RMSE & MAPE(\%) \\ \hline
\multirow{12}{*}{Alameda} 
 & HL & 15.60 & 26.44 & 18.62 & 18.48 & 31.50 & 21.61 & 24.86 & 41.63 & 26.98 & 19.06 & 32.24 & 21.72 \\
 & LSTM & 13.88 & 23.15 & 17.34 & 15.86 & 26.86 & 20.00& 19.79 & 32.98 & 22.47 & 16.04 & 26.82 & 19.83 \\
 & DCRNN & 13.63 & 23.22 & 17.90 & 15.68 & 27.01 & 19.83 & 19.53 & 32.90 & 22.39 & 15.78 & 26.83 & 19.49 \\
 & AGCRN & 13.07 & 22.01 & 17.81 & 14.70 & 25.01 & 19.46 & 16.81 & 27.79 & 22.47 & 14.34 & 24.03 & 19.78 \\
 & STGCN & 15.73 & 25.30 & 22.26 & 17.68 & 28.88 & 28.14 & 21.31 & 35.43 & 29.70 & 17.79 & 29.09 & 25.65 \\
 & GWNET & 14.87 & 24.48 & 18.93 & 17.69 & 29.49 & 21.45 & 22.51 & 37.29 & 31.44 & 17.76 & 29.49 & 22.77 \\
 & ASTGCN & 14.60 & 24.69 & 17.00 & 16.79 & 28.86 & 21.26 & 21.86 & 36.66 & 30.79 & 17.16 & 29.08 & 22.08 \\
 & STTN & 13.39 & 22.07 & 18.00 & 14.89 & 24.90 & 20.20 & 17.86 & 29.36 & 22.25 & 15.08 & 24.98 & 20.16 \\
 & DSTAGNN & \underline{12.48} & \underline{21.25} & 14.74 & \underline{13.53} & \underline{23.13} & \underline{15.62} & \underline{15.24} & \underline{25.27} & 30.10 & \underline{13.45} & \underline{22.63} & 19.28 \\ 
 & DGCRN & 14.18 & 23.77 & 20.24 & 16.79 & 28.72 & 24.17 & 24.44 & 39.97 & 40.04 & 17.76 & 29.71 & 26.81 \\
 & D$^2$STGNN & 13.19 & 22.41 & 15.40 & 14.64 & 24.90 & 16.65 & 17.09 & 28.39 & 19.91 & 14.62 & 24.67 & 17.09 \\
 & BiST & 13.33 & 23.05 & \underline{14.71} & 14.28 & 24.56 & 17.42 & 16.60 & 27.98 & \underline{19.30} & 14.27 & 24.17 & \underline{16.60} \\
 \cline{2-14} 
 & IGSTGNN & \textbf{11.80} & \textbf{20.08} & \textbf{14.33} & \textbf{12.64} & \textbf{21.70} & \textbf{15.22} & \textbf{14.21} & \textbf{24.37} & \textbf{16.97} & \textbf{12.69} & \textbf{21.73} & \textbf{15.43} \\ \midrule
\multirow{12}{*}{\shortstack{Contra\\ Costa}}
 & HL & 16.47 & 27.85 & 16.58 & 18.64 & 31.28 & 20.65 & 23.44 & 37.52 & 24.65 & 19.15 & 31.71 & 19.95 \\
 & LSTM & 14.78 & 24.95 & 15.63 & 16.89 & 29.02 & 19.65 & 20.04 & 32.74 & 19.73 & 16.88 & 28.31 & 17.74 \\
 & DCRNN & 14.42 & 24.81 & 14.95 & 16.00 & 28.06 & 18.37 & 19.25 & 31.90 & 18.87 & 16.17 & 27.67 & 16.50 \\
 & AGCRN & 14.65 & 24.36 & 14.96 & 15.91 & 27.06 & 18.78 & 18.41 & 29.38 & 19.18 & 15.98 & 26.51 & 16.94 \\
 & STGCN & 16.39 & 24.66 & \textbf{13.85} & 20.75 & 30.82 & \textbf{14.93} & 23.12 & 33.77 & 19.85 & 19.71 & 29.35 & \textbf{15.46} \\
 & GWNET & 14.69 & 24.67 & 17.16 & 15.93 & 27.78 & 20.16 & 19.16 & 30.89 & 19.21 & 16.46 & 27.50 & 18.24 \\
 & ASTGCN & 14.94 & 25.38 & 15.75 & 17.15 & 29.25 & 22.72 & 20.44 & 33.00 & 27.42 & 17.31 & 28.68 & 20.93 \\
 & STTN & 14.23 & 23.95 & 15.48 & 15.32 & 26.14 & 19.00 & 18.32 & 29.06 & 19.95 & 15.59 & 26.02 & 17.19 \\
 & DSTAGNN & 14.35 & 24.51 & 14.75 & 15.48 & 26.32 & 19.02 & 16.56 & 27.19 & \textbf{16.83} & 15.28 & 25.74 & 16.65 \\
 & DGCRN & 15.62 & 25.32 & 18.44 & 19.38 & 30.73 & 29.05 & 29.01 & 43.17 & 42.05 & 20.39 & 31.70 & 28.00 \\ 
 & D$^2$STGNN & \underline{12.91} & \underline{21.76} & 17.26 & \underline{14.03} & \underline{23.47} & 20.45 & \underline{16.01} & \underline{25.70} & 23.89 & \underline{14.09} & \underline{23.31} & 20.19 \\
 & BiST & 13.31 & 23.29 & 14.57 & 14.67 & 25.88 & 19.16 & 16.59 & 27.12 & 18.04 & 14.62 & 25.13 & 16.18 \\
 \cline{2-14} 
 & IGSTGNN & \textbf{12.50} & \textbf{21.37} & \underline{14.14} & \textbf{13.44} & \textbf{22.83} & \underline{16.01} & \textbf{14.81} & \textbf{24.13} & \underline{17.95} & \textbf{13.43} & \textbf{22.50} & \underline{15.84} \\ \midrule
\multirow{12}{*}{Orange}
 & HL & 16.79 & 28.75 & 17.35 & 20.52 & 34.87 & 20.56 & 25.76 & 42.49 & 24.77 & 20.51 & 34.59 & 20.34 \\
 & LSTM & 14.47 & 24.70 & 15.43 & 16.94 & 29.53 & 17.63 & 20.66 & 34.91 & 19.61 & 17.02 & 29.24 & 17.29 \\
 & DCRNN & 14.07 & 24.22 & \underline{14.24} & 16.11 & 27.73 & 15.92 & 19.11 & 32.57 & 18.00 & 16.11 & 27.68 & 15.95 \\
 & AGCRN & 14.48 & 23.44 & 20.74 & 16.48 & 27.73 & 19.84 & 18.34 & 30.79 & 22.61 & 16.20 & 27.19 & 20.79 \\
 & STGCN & 15.82 & 24.81 & 20.07 & 18.22 & 29.61 & 21.35 & 19.76 & 32.31 & 21.22 & 17.79 & 28.99 & 20.82 \\
 & GWNET & 14.69 & 24.70 & 16.33 & 17.14 & 29.77 & 19.07 & 19.95 & 33.81 & 21.12 & 16.99 & 29.01 & 18.40 \\
 & ASTGCN & 14.71 & 25.50 & 15.65 & 17.66 & 31.10 & 18.63 & 21.83 & 37.06 & 24.12 & 17.56 & 30.49 & 18.89 \\
 & STTN & 13.09 & \underline{22.54} & 14.54 & \underline{13.91} & \underline{24.01} & \underline{15.61} & \underline{14.96} & \underline{25.74} & 16.60 & \underline{13.82} & \underline{23.80} & \underline{15.39} \\
 & DSTAGNN & 13.69 & 22.97 & 19.44 & 14.36 & 25.03 & 16.02 & 15.69 & 27.03 & \textbf{16.06} & 14.61 & 25.03 & 16.51 \\
 & DGCRN & 13.99 & 23.53 & 17.75 & 16.00 & 27.01 & 22.81 & 19.89 & 31.77 & 33.36 & 16.14 & 26.84 & 23.25 \\ 
 & D$^2$STGNN & 13.11 & 22.83 & 15.68 & 13.93 & 24.27 & 17.28 & 15.31 & 26.52 & 18.02 & 13.92 & 24.21 & 16.83 \\
 & BiST & \underline{13.04} & 22.68 & 15.76 & 14.65 & 25.37 & 17.37 & 16.75 & 29.16 & 19.72 & 14.56 & 25.39 & 17.23 \\
 \cline{2-14} 
 & IGSTGNN & \textbf{12.28} & \textbf{21.95} & \textbf{13.82} & \textbf{13.16} & \textbf{23.45} & \textbf{15.16} & \textbf{14.35} & \textbf{25.48} & \underline{16.10} & \textbf{13.13} & \textbf{23.35} & \textbf{14.80} \\ 
\bottomrule
\end{tabular}
}
\end{table*}

\subsubsection{Datasets}

Our experimental data are sourced from the large-scale XTraffic~\cite{gou2024xtraffic} benchmark dataset, which contains comprehensive traffic flow and incident records from California for the year 2023. To focus on freeway mainline traffic dynamics, we exclusively select data from mainline sensors. All data is aggregated into 5-minute time windows. We constructed three distinct sub-datasets for our evaluation, corresponding to three different regions in California: Alameda, Contra Costa, and Orange. Each dataset is partitioned into training, validation, and testing sets with a 70\%/15\%/15\% ratio. Our task is formulated to predict the traffic conditions for the next 1 hour (12 steps) using the historical sequence from the past 1 hour (12 steps). Detailed statistics of the datasets are presented in Table~\ref{tab:dataset_stats}.

\textit{\textbf{Feature Details.}} To fully utilize the incident context, we extracted rich attributes from the raw logs. The specific \textbf{Sensor Meta-features} (e.g., road type, speed limit) and \textbf{Incident Information Features} (e.g., type, description, location) used in our model are detailed in Table~\ref{tab:sensor_features} and Table~\ref{tab:incident_features}, respectively.

\subsubsection{Baselines on Benchmarks}

We compare our proposed IGSTGNN with a series of state-of-the-art spatio-temporal forecasting models. To ensure a fair comparison, our forecasting experiments were implemented within the same software framework employed by LargeST~\cite{liu2023largest}, and we further adhered to the identical experimental settings outlined within that work. 
The selected baselines include classic methods (HL~\cite{liang2020revisiting}, LSTM~\cite{hochreiter1997long}) and a variety of spatio-temporal graph models. The latter can be further categorized by their temporal modeling approach into: \textbf{RNN-based} (DCRNN~\cite{li2017diffusion}, AGCRN~\cite{bai2020adaptive}), \textbf{CNN-based} (STGCN~\cite{Yu2017Spatio-temporal}, GWNET~\cite{wu2019graph}), \textbf{attention-based} (ASTGCN~\cite{guo2019attention}, STTN~\cite{xu2020spatial}), and recent \textbf{dynamic graph-based} approaches (DSTAGNN~\cite{lan2022dstagnn}, DGCRN~\cite{li2023dynamic}, D$^2$STGNN~\cite{shao2022decoupled}, BiST~\cite{ma2025bist}).

\subsubsection{Implementation Details}

For our proposed IGSTGNN model, we use the Adam optimizer with an initial learning rate of 0.002 and an early stopping strategy (patience of 20 epochs) employed to prevent overfitting. The batch size was set to 48 for the Alameda and Contra Costa datasets, and adjusted to 24 for the larger Orange dataset. Key architectural hyperparameters were configured as follows: in the feature encoding stage, we construct 8-dimensional and 32-dimensional embeddings for the 6 incident types and 30 incident descriptions, respectively; the ST modeling module is composed of 5 decoupling blocks, where the dimension for both adaptive node and timestamp embeddings is set to 12; and in the \textit{TIID} module, the temporal decay parameter $\sigma_t$ is set to 1.0. All experiments were conducted on a server equipped with an Intel Xeon Gold 6138 CPU and an NVIDIA Tesla V100 (32GB) GPU\footnote{Code is available at \url{https://github.com/fanlixiang/IGSTGNN}.}.

\subsubsection{Evaluation Metrics}

Model performance is evaluated using three standard metrics: Mean Absolute Error (MAE), Root Mean Squared Error (RMSE), and Mean Absolute Percentage Error (MAPE), where lower values indicate better performance.

\subsection{Performance Comparison with Baselines}

We compare our proposed IGSTGNN model against all baselines on the three datasets to evaluate its overall effectiveness. The detailed quantitative results are summarized in Table~\ref{tab:main_results_all_datasets}, which reports the performance over three prediction horizons (15, 30, and 60 minutes ahead) as well as the average performance.

The results clearly demonstrate the superiority of our proposed IGSTGNN framework, which consistently outperforms all state-of-the-art baselines across all datasets and prediction horizons. For instance, on the Alameda dataset, IGSTGNN achieves an average MAE that is \textbf{5.65\%} lower than the second-best model. This consistent advantage strongly validates the effectiveness of our incident-aware design. Key observations are as follows:

\textbf{Robustness in Long-Term Forecasting}. IGSTGNN consistently maintains its leading position in the most challenging long-term forecasting tasks. For example, on the Orange dataset, IGSTGNN still maintains a significant 4.1\% advantage in MAE over the strong second-best model, STTN, at Horizon 12. This directly validates the effectiveness of our \textit{TIID} module in capturing the dynamics of traffic recovery. This robustness is particularly important for real-world deployment, where operators rely on stable forecasts under rare but severe incidents.

\textbf{Adaptability to Spatial Heterogeneity}. IGSTGNN maintains its leading performance despite the varying network structures and incident distributions across the three datasets. This suggests that our \textit{ICSF} module can effectively handle the spatial heterogeneity of incident impacts by generating a customized incident impact representation for each node.

\subsection{Effectiveness of Proposed Modules}

To validate the generalizability of our proposed \textit{ICSF} and \textit{TIID} modules as "plug-and-play" components, we conducted a module effectiveness study. To ensure a comprehensive evaluation, we selected several state-of-the-art STGNNs from our baselines that represent distinct architectural paradigms: an RNN-based model (AGCRN), a CNN-based model (GWNET), a Transformer-based model (STTN), and a hybrid model employing a spatio-temporal decoupling framework (D$^2$STGNN). We integrated the \textit{ICSF} and \textit{TIID} modules with these models in three settings: \textit{ICSF} only, \textit{TIID} only, and both modules combined, evaluating their performance changes across different datasets.

\begin{figure}[!ht]
    \centering
    \includegraphics[width=\columnwidth]{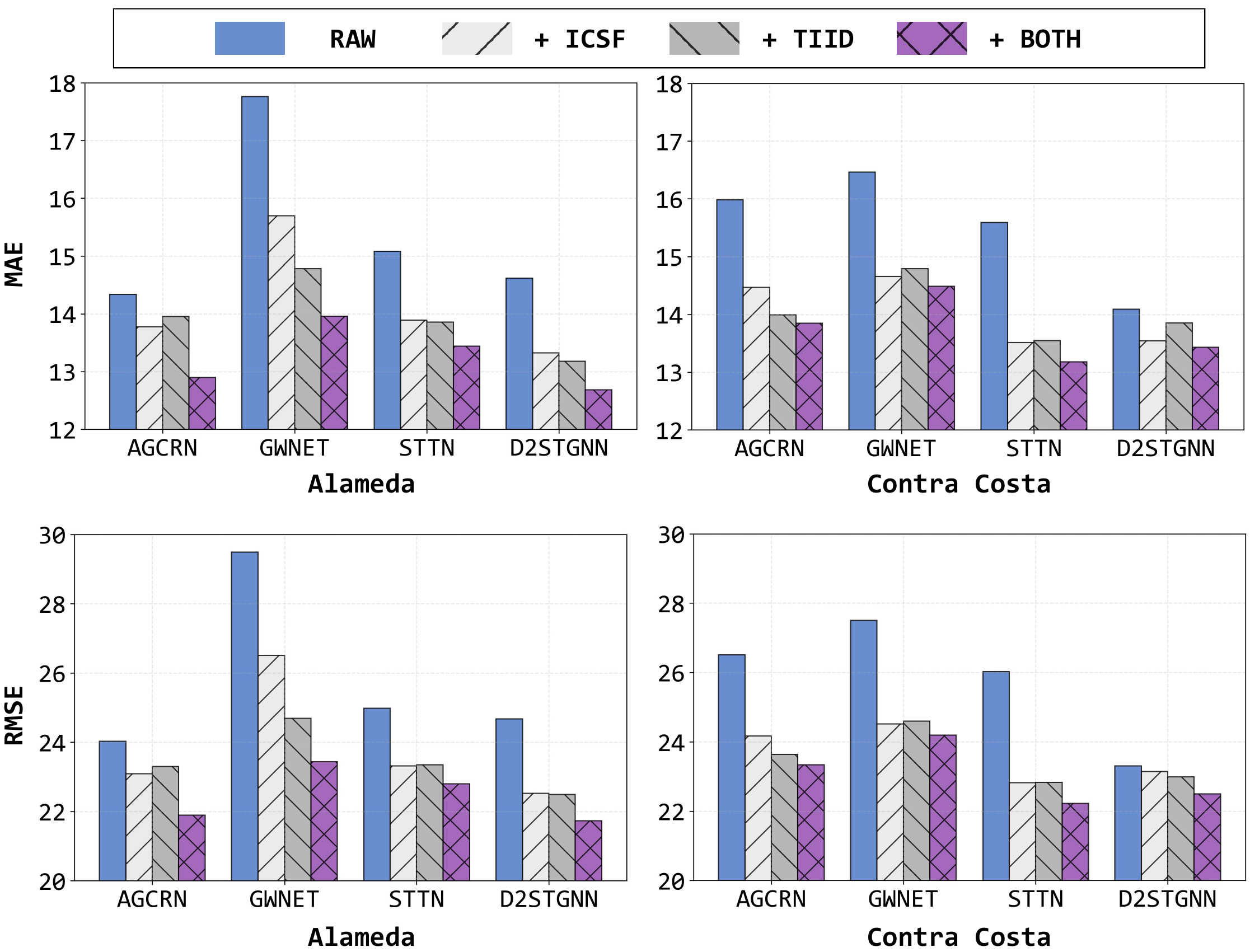} 
    \caption{Effectiveness of the \textit{ICSF} and \textit{TIID} modules. The figure compares the performance (MAE and RMSE) of baseline models in their original form (Raw) versus after integrating the modules separately (\textit{ICSF}, \textit{TIID}) and in combination (BOTH) on the Alameda and Contra Costa datasets.}
    \Description{
        Bar charts demonstrating the performance improvements from the proposed modules. 
        The x-axis represents different baseline models. 
        The y-axis represents error metrics (MAE and RMSE). 
        Grouped bars show four settings: Raw (original), +\textit{ICSF}, +\textit{TIID}, and Both. 
        The 'Both' configuration consistently shows the lowest bar height, indicating the best performance (lowest error) across Alameda and Contra Costa datasets.
        }
    \label{fig:effectiveness_chart}
\end{figure}

The results are visualized in Figure~\ref{fig:effectiveness_chart}. It is clearly observable that while integrating either the \textit{ICSF} or \textit{TIID} module alone brings consistent performance gains to all tested baselines, combining both modules achieves the most significant improvements, demonstrating a strong synergistic effect. For instance, on the Alameda dataset, integrating the \textit{ICSF} and \textit{TIID} modules with the D$^2$STGNN model separately yields MAE improvements of \textbf{8.85\%} and \textbf{9.85\%}, respectively. When both modules were enabled concurrently, the performance improvement soared to \textbf{13.23\%}, proving their functional complementarity. This trend is highly consistent across diverse model architectures and datasets. These findings strongly demonstrate that our proposed \textit{ICSF} and \textit{TIID} modules are not only effective on their own but also work synergistically as general, "plug-and-play" enhancement components. They can empower other spatio-temporal forecasting models with the crucial capabilities of contextual fusion and temporal decay modeling required for handling traffic incidents.

\subsection{Superiority of the \textit{ICSF} Module}

Finally, to demonstrate the superiority of our designed \textit{ICSF} module compared to simpler fusion methods, we conduct a comparative experiment. In this experiment, we replace the \textit{ICSF} module within the full IGSTGNN framework with two alternatives: The first is \textbf{MLP}, which simply concatenates the embedding vectors of all information and fuses them through an MLP. The second is \textbf{Iterative Message Passing (IMP)}, which employs a more complex, message-passing-like mechanism where incident and node representations undergo multiple rounds of interactive updates.

\begin{figure}[!t]
    \centering
    \includegraphics[width=\columnwidth]{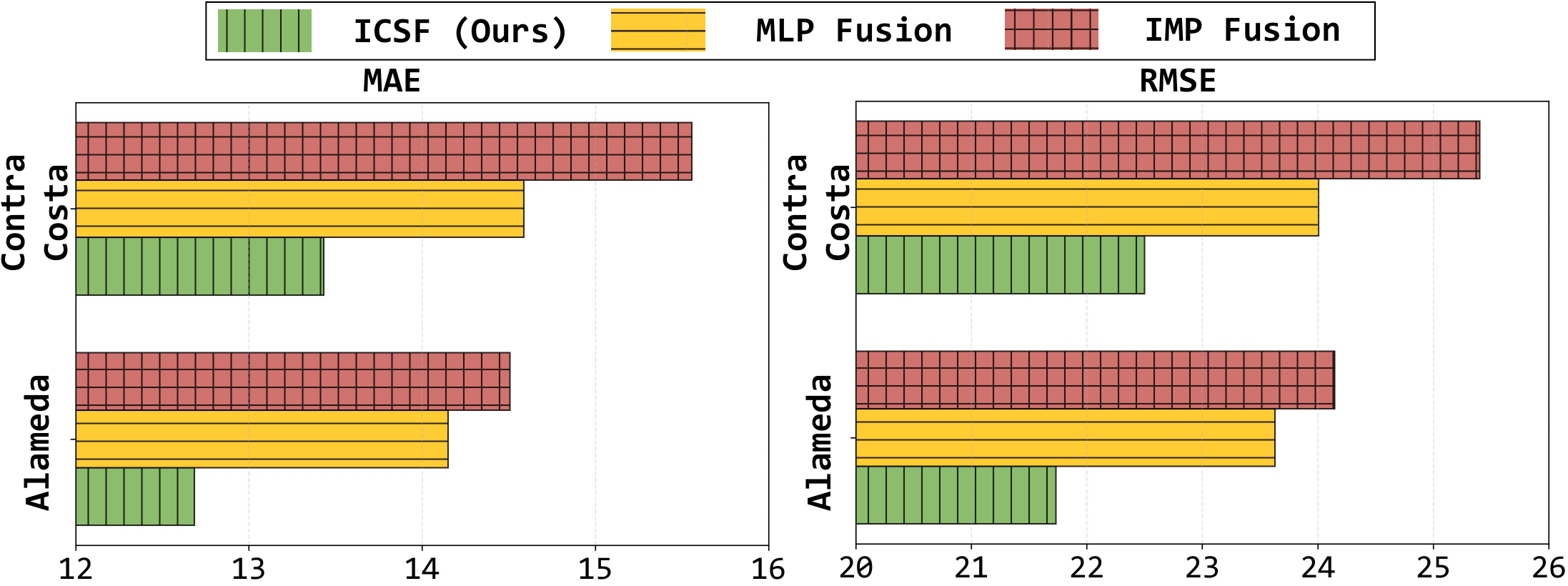} 
    \caption{Performance comparison (in terms of average MAE and RMSE) of our \textit{ICSF} module against MLP and IMP fusion methods on the Alameda and Contra Costa datasets.}
    \Description{
    A performance comparison chart.
    The chart contrasts the proposed \textit{ICSF} module against two other fusion methods: MLP and IMP.
    The vertical axis represents error values (MAE and RMSE).
    The bars show that \textit{ICSF} consistently yields lower error rates compared to MLP and IMP on both Alameda and Contra Costa datasets, demonstrating its superiority.
    }
    \label{fig:superiority_chart}
\end{figure}

As shown in Figure~\ref{fig:superiority_chart}, our proposed context-aware attention-based \textit{ICSF} module significantly outperforms both the MLP and IMP alternatives in terms of average MAE and RMSE on both datasets. Specifically, on the Alameda dataset, the MAE of \textit{ICSF} is \textbf{10.3\%} lower than that of MLP and \textbf{12.5\%} lower than that of IMP. This result validates our design choice: simple feature concatenation (MLP) is insufficient to capture the complex non-linear relationships between incidents, sensors, and traffic states, while overly complex interactions (IMP) may be difficult to optimize. In contrast, the attention mechanism in \textit{ICSF} strikes a better balance between capturing critical information and modeling complex relationships, thereby achieving superior performance.

\subsection{Ablation Study}

To deeply investigate the effectiveness of our model design, we conduct a series of detailed ablation studies. We analyze the contributions from two perspectives: (1) \textbf{Module Ablation}, where we remove the \textit{ICSF} or \textit{TIID} modules respectively, and (2) \textbf{Feature Ablation}, where we remove sensor meta-features, incident intrinsic attributes, or distance information.

\begin{table}[!htbp]
\centering
\caption{Ablation study of IGSTGNN and its variants on Alameda, Contra Costa, and Orange datasets. All metrics are averaged over the 12-step prediction horizon.}
\label{tab:ablation_study}
\resizebox{\columnwidth}{!}{
\begin{tabular}{l l|ccc}
\toprule
\multirow{2}{*}{Data} & \multirow{2}{*}{Method} & \multicolumn{3}{c}{Average} \\ \cline{3-5} 
 &  & MAE & RMSE & MAPE(\%) \\
\midrule
\multirow{6}{*}{Alameda} 
 & w/o \textit{ICSF} & 13.23 & 22.68 & 17.27 \\
 & w/o \textit{TIID} & 13.23 & 22.64 & \textbf{15.13} \\
 & w/o S (sensor attr.) & 13.21 & 22.61 & 15.44 \\
 & w/o D (distance) & 13.20 & 22.61 & 18.44 \\
 & w/o I (incident attr.) & 13.09 & 22.48 & 15.96 \\
 \cline{2-5}
 & IGSTGNN (Ours) & \textbf{12.69} & \textbf{21.73} & 15.43 \\
\midrule
\multirow{6}{*}{Contra Costa} 
 & w/o \textit{ICSF} & 13.52 & 23.25 & 16.76 \\
 & w/o \textit{TIID} & 13.54 & 22.83 & 17.35 \\
 & w/o S (sensor attr.) & 13.52 & 23.08 & 16.14 \\
 & w/o D (distance) & 13.55 & 23.24 & 17.16 \\
 & w/o I (incident attr.) & 13.59 & 22.99 & \textbf{15.80} \\
 \cline{2-5}
 & IGSTGNN (Ours) & \textbf{13.43} & \textbf{22.50} & 15.84 \\
\midrule
\multirow{6}{*}{Orange} 
 & w/o \textit{ICSF} & 13.38 & 23.60 & 14.84 \\
 & w/o \textit{TIID} & 13.32 & 23.57 & 14.82 \\
 & w/o S (sensor attr.) & 13.45 & 23.68 & \textbf{14.53} \\
 & w/o D (distance) & 13.60 & 23.72 & 17.50 \\
 & w/o I (incident attr.) & 13.59 & 23.66 & 15.55 \\
 \cline{2-5}
 & IGSTGNN (Ours) & \textbf{13.13} & \textbf{23.35} & 14.80 \\
\bottomrule
\end{tabular}
}
\end{table}

Table~\ref{tab:ablation_study} presents the quantitative results of the ablation study across three datasets. Specifically, we removed each component or feature input to isolate its contribution. The results consistently show that the full IGSTGNN achieves the best MAE/RMSE across all three datasets, while MAPE is largely comparable with only minor variations in a few cases, supporting the overall effectiveness of the proposed framework. We draw three key observations:

First, both the \textit{ICSF} and \textit{TIID} modules contribute consistently to performance. Removing either module increases MAE and RMSE across all datasets, though the magnitude varies by dataset. For instance, on Alameda, removing \textit{ICSF} or \textit{TIID} increases the MAE to 13.23 (a 4.2\% degradation compared to the full model), indicating that modeling both the initial spatial heterogeneity and the temporal decay of incident impacts is important.

Second, incorporating \textbf{incident attributes (I)} provides complementary gains. The variant ``w/o I'' (removing incident type and description inputs) shows a degradation (e.g., MAE increases from 13.13 to 13.59 on Orange), suggesting that incident semantics offer additional discriminative cues beyond spatial proximity and sensor context.

Finally, spatial context features, including \textbf{Distance (D)} and \textbf{Sensor Attributes (S)}, also play an important role. In particular, removing distance information leads to a notable degradation (e.g., MAPE rises to 17.50 on Orange), confirming that precise spatial anchoring and road network context are crucial for modeling how incident impacts propagate through the traffic graph.

\section{Related Work}
\label{sec:related_work}

This section reviews the relevant literature from two perspectives: the mainstream models for spatio-temporal traffic forecasting and related works about the impact of external incidents on traffic.

\subsection{Spatio-Temporal Traffic Forecasting}
Spatio-temporal traffic forecasting has evolved from traditional statistical methods (e.g., ARIMA) and early deep learning models (e.g., LSTM), which have limitations in capturing complex spatio-temporal dependencies~\cite{Yu2017Spatio-temporal, Zhang2019Short-term, Dai2020Spatio-Temporal, Pan2019Urban}. Currently, the dominant paradigm involves combining Graph Neural Networks with various temporal modeling components. The core idea is to abstract the traffic network as a graph, utilizing Graph Neural Networks (GNNs) to capture spatial dependencies while combining them with Recurrent Neural Networks (RNNs), Convolutional Neural Networks (CNNs), or attention mechanisms to model temporal dynamics~\cite{Yu2017Spatio-temporal, Dai2020Spatio-Temporal, Han2021Dynamic, Du2024Multi-scale, Fang2022Learning, Wu2024Multi-Scale, 2022Spatio‐temporal, Ta2022Adaptive, Jiang2022Spatio-Temporal, DBLP:conf/icde/CirsteaYGKP22}. 

Recent works have further explored decoupling complex spatio-temporal patterns and adaptively learning the graph structure~\cite{Ta2022Adaptive, Jiang2022Spatio-Temporal}. Based on their temporal modeling approach, these advanced models can be categorized into: \textbf{RNN-based} (DCRNN~\cite{li2017diffusion}, AGCRN~\cite{bai2020adaptive}), \textbf{CNN-based} (STGCN~\cite{Yu2017Spatio-temporal}, GWNET~\cite{wu2019graph}), \textbf{attention-based} (ASTGCN~\cite{guo2019attention}, STTN~\cite{xu2020spatial}), and recent \textbf{dynamic graph-based} approaches (DSTAGNN~\cite{lan2022dstagnn}, DGCRN~\cite{li2023dynamic}, D$^2$STGNN~\cite{shao2022decoupled}, BiST~\cite{ma2025bist}). Despite the great success of these graph-based deep learning models in learning from historical traffic data, their core design is oriented towards regular, periodic traffic patterns, which limits their ability to model the atypical dynamics induced by external, abrupt incidents.

\subsection{The Incident Impact in Traffic Analysis}

Incorporating external information has become a key research direction for improving traffic forecasting accuracy, especially when purely historical signals fail to explain abrupt state transitions~\cite{Ahmed2024Enhancement, Zhu2020KST-GCN, Zhou2023Integrating, Zhu2020AST-GCN, Abadi2015Traffic}. 
Among various external sources, traffic incidents (e.g., accidents, road works, and control measures) have attracted significant attention due to their strong association with congestion and sudden changes in traffic states~\cite{Zhu2020AST-GCN, Ahmed2024Enhancement, Abadi2015Traffic}. 
Moreover, the multi-dimensional attributes of incidents---such as type, location, and duration---provide rich, quantifiable features for modeling traffic flow variations~\cite{Zhu2020AST-GCN, Zhu2020KST-GCN, Zhou2023Integrating}.

More recently, specialized incident-aware models have been proposed to better handle the irregularity and sparsity of incident patterns. 
For example, DIGC-Net introduces a two-stage critical incident discovery and impact representation learning pipeline, which identifies critical incidents and aggregates sequential incident information within a time window before fusing it with spatio-temporal predictors~\cite{xie2020deep}. 
In parallel, broader event-aware frameworks (beyond road incidents) have been explored for mobility forecasting under societal events such as holidays, severe weather, and epidemics; EAST-Net leverages memory-augmented dynamic parameter generation to enhance adaptivity under unprecedented events~\cite{wang2022event}.

Despite these advances, incident effects are often incorporated through implicit feature fusion (e.g., concatenation/attention/sequence aggregation) rather than being explicitly formulated as a spatio-temporal propagation and attenuation process on the traffic network. 
As a result, existing methods may still be limited in capturing (i) heterogeneous spatial spread of localized disruptions and (ii) the continuous temporal evolution and decay of incident impacts.

\section{Conclusion}
\label{sec:conclusion}

In this paper, we proposed IGSTGNN, a novel framework designed to address the challenges of forecasting in incident-affected scenarios. To systematically model an incident's influence, the framework utilized two core components: an Incident-Context Spatial Fusion (\textit{ICSF}) module to capture the non-uniform spatial impact, and a Temporal Incident Impact Decay (\textit{TIID}) module to model the dynamic temporal evolution of the impact. Comprehensive experiments on multiple large-scale, real-world datasets demonstrated that the performance of IGSTGNN was significantly superior to state-of-the-art baseline models. Our work provides an effective and systematic solution for incident-guided spatio-temporal forecasting. To support reproducibility, we release the dataset to facilitate future research and fair comparisons. Moreover, the proposed \textit{ICSF} and \textit{TIID} modules can be integrated as plug-and-play components, consistently improving several STGNN backbones. Future work could explore extending the framework to incorporate a broader range of external disturbances, and leveraging foundation models or agentic pipelines to extract and normalize incident semantics.

\begin{acks}
This work was supported by the National Natural Science Foundation of China under Grant No. U2469205, and the Fundamental Research Funds for the Central Universities of China under Grant No. JKF-20240769.
\end{acks}

\clearpage
\balance
\bibliographystyle{ACM-Reference-Format}
\bibliography{reference}

\clearpage

\appendix
\section{Appendix}

\subsection{Data Processing Details}

The construction of the final dataset involved a multi-stage pipeline, ensuring precise alignment between traffic time series and incident logs:
\begin{itemize}
    \item \textbf{Data Filtering and Selection:} We began by loading the raw sensor metadata and incident logs. To create focused and manageable datasets, we filtered this data to include only mainline sensors from three specific California counties (Alameda, Contra Costa, and Orange) for the year 2023.
    \item \textbf{Temporal Alignment:} The continuous traffic sensor readings were aggregated into discrete 5-minute time windows. Each incident record was then precisely aligned and mapped to a single 5-minute window based on its occurrence timestamp. This step is crucial for synchronizing the incident information with the traffic time series.
    \item \textbf{Geocoding and Spatial Relation Tensor Construction:} To compute spatial relationships, a unified coordinate system is required. While sensors had latitude and longitude, incidents were located by postmiles. We used the official \textbf{Caltrans Postmile Query Tool} to convert incident postmiles into precise latitude and longitude coordinates. With this information, we constructed the spatial relationship tensor ($\bm{D}$), whose three dimensions for each incident-sensor pair represent their Gaussian-transformed Euclidean distance, Gaussian-transformed road network distance, and their binary upstream/downstream relationship.
    \item \textbf{Feature Engineering and Data Splitting:} Additional time-based features, such as the time of day and day of the week, were engineered from the timestamps. Finally, the complete time-ordered dataset for each region was chronologically split into training (70\%), validation (15\%), and testing (15\%) sets.
\end{itemize}

\subsection{Incident Taxonomy}
To clarify the scope of external disturbances modeled in this study, we list the specific incident types and descriptions included in our dataset in Table~\ref{tab:incident_definitions}. The dataset covers 6 major categories and 30 fine-grained descriptions.

\begin{table}[!h]
\centering
\caption{Taxonomy and Definitions of Incident Types.}
\label{tab:incident_definitions}
\resizebox{\columnwidth}{!}{%
\begin{tabular}{l | p{3.8cm} | p{5.2cm}}
\toprule
\textbf{Incident Type} & \textbf{Typical Examples} & \textbf{Key Definitions / Clarification} \\
\midrule
\textbf{Hazard} & Traffic Hazard, Debris & Impediments obstructing traffic flow (e.g., objects/animals). \\ \hline
\textbf{Accident} & 1141 En Route, Collision & Vehicle collisions. \textbf{``1141''}: Injury accident requiring ambulance response. \\ \hline
\textbf{Breakdown} & Disabled Vehicle, Flat Tire & Vehicles stopped due to mechanical failures (non-collision). \\ \hline
\textbf{Weather} & Fog, Wind, Rain, Snow & Adverse atmospheric conditions affecting visibility/friction. \\ \hline
\textbf{Other} & Fire, Sigalert, Roadwork & Misc. events. \textbf{``Fire''}: Vehicle or roadside fire. \\ \hline
\textbf{Police} & Police Activity, Advisory & Law enforcement activities not directly involving collisions. \\
\bottomrule
\end{tabular}%
}
\end{table}

\subsection{Supplementary Experimental Results}

\subsubsection{Performance by Incident Type}
To evaluate the generalization capability of IGSTGNN across different disturbances, we analyzed the model's performance on subsets of test data filtered by incident type. As shown in Table~\ref{tab:type_performance}, IGSTGNN consistently achieves low prediction errors across diverse categories, outperforming baselines even when they are enhanced with incident features.

\begin{table}[h]
\centering
\caption{Performance comparison (MAE) across different incident types on the Alameda dataset.}
\label{tab:type_performance}
\resizebox{\columnwidth}{!}{
\begin{tabular}{l|c|cc}
\toprule
\textbf{Incident Type} & \textbf{IGSTGNN (Ours)} & \textbf{DSTAGNN+I} & \textbf{D$^2$STGNN+I} \\
\midrule
Hazard & \textbf{12.82} & 13.02 & 13.37 \\
Accident & \textbf{13.22} & 13.52 & 14.17 \\
Breakdown & \textbf{12.40} & 13.01 & 12.94 \\
Weather & \textbf{8.50} & 9.36 & 10.56 \\
Other & \textbf{14.25} & 14.47 & 14.98 \\
Police & \textbf{15.11} & 15.62 & 16.22 \\
\bottomrule
\end{tabular}
}
\end{table}

\subsubsection{\textit{ICSF} Module Superiority Study}
In this section, we compare \textit{ICSF} with two representative baseline fusion approaches, which are detailed below.

\paragraph{MLP Fusion} This method serves as a simple fusion baseline. It simply \textbf{concatenates} all relevant feature vectors (e.g., incident and sensor representations) into a single, flattened vector. This combined vector is then processed by a Multi-Layer Perceptron (MLP) to produce a fused output, without explicitly modeling any structural or spatial relationships.

\paragraph{IMP Fusion (Iterative Message Passing)} This is a more complex, graph-inspired approach. In this mechanism, incident and sensor nodes are treated as nodes in a bipartite graph. For a fixed number of iterations, they exchange information through a \textbf{message-passing} process, where each node updates its representation based on the aggregated messages from its neighbors in the previous round.

\begin{figure}[ht]
    \centering
    \includegraphics[width=0.7\columnwidth]{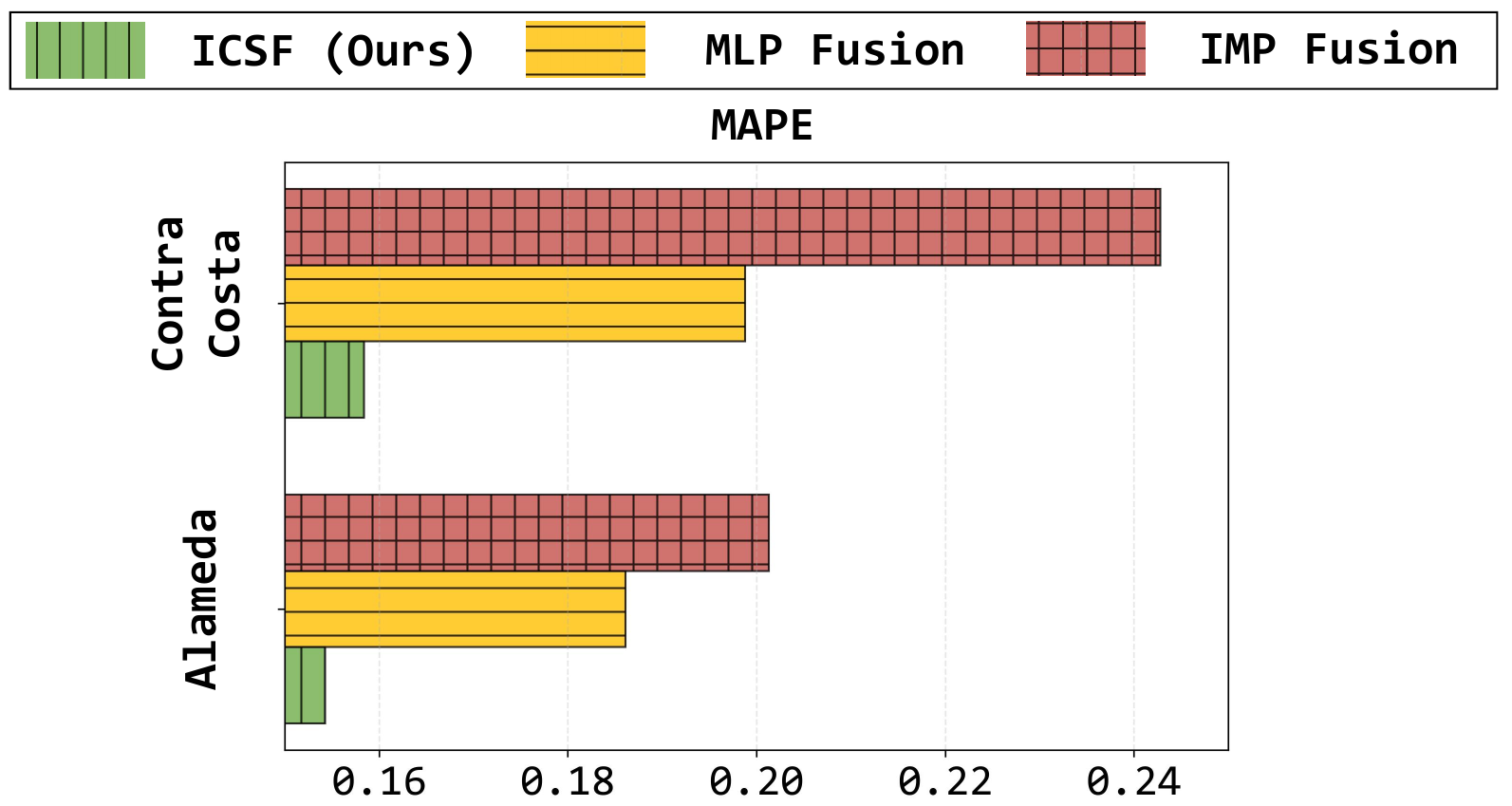}
    \caption{MAPE (\%) results of the \textit{ICSF} module superiority study.}    
    \Description{
    A bar chart illustrating MAPE percentage errors for the superiority study.
    It compares the proposed \textit{ICSF} module against MLP Fusion and IMP Fusion methods.
    The bars for the \textit{ICSF} module are lower than those for MLP and IMP, indicating superior performance.
    }
    \label{fig:superiority_appendix}
\end{figure}

The MAPE results, shown in Figure~\ref{fig:superiority_appendix}, demonstrate the superiority of our proposed \textit{ICSF} module. On both datasets, \textit{ICSF} achieves the lowest MAPE. This result validates our design choice: simple feature concatenation (MLP) is insufficient to capture complex spatial relationships, while an overly complex iterative method (IMP) may be difficult to optimize. In contrast, our \textit{ICSF} module strikes a more effective balance between fusion effectiveness and model complexity.

\balance

\end{document}